# Machine Learning Approaches for Amharic Parts-of-speech Tagging

**Ibrahim Gashaw**
Mangalore University
Mangalagangotri, Mangalore-574199
`ibrahimug1@gmail.com`

**H L Shashirekha**
Mangalore University
Mangalagangotri, Mangalore-574199
`hlsrekha@gmail.com`

## Abstract

Part-of-speech (POS) tagging is considered as one of the basic but necessary tools which are required for many Natural Language Processing (NLP) applications such as word sense disambiguation, information retrieval, information processing, parsing, question answering, and machine translation. Performance of the current POS taggers in Amharic is not as good as that of the contemporary POS taggers available for English and other European languages. The aim of this work is to improve POS tagging performance for the Amharic language, which was never above 91%. Usage of morphological knowledge, an extension of the existing annotated data, feature extraction, parameter tuning by applying grid search and the tagging algorithms have been examined and obtained significant performance difference from the previous works. We have used three different datasets for POS experiments.

## 1 Introduction

POS tagging is the process of assigning the part of speech categories to each and every word in a sentence. In many NLP applications such as word sense disambiguation, information retrieval, information processing, parsing, question answering, and machine translation, it is considered as one of the basic but necessary tool that could be utilized in computational linguistics analysis and automation applications (Antony and Soman, 2011).

Existing POS tagger approaches can be classified into: linguistic (rule-based), statistical/machine-learning and hybrid approaches.

**Linguistic approachs:** Most POS taggers arrange linguistic knowledge systematically as a set of rules (or constraints) written by linguists that range from a few hundred to several thousand, and usually require a high cost for experts and consume time (Màrquez et al., 2000).

**Statistical/Machine Learning approaches:** These approaches use frequency or probability to tag words in a text. With the simplest Statistical tagger, the ambiguity of words established on the probability that the word occurs alongside a particular tag can be resolved. Statistical approach involves some kind of learning (supervised or unsupervised) parameters of the model from a training corpus (Radziszewski, 2013).

**Hybrid approaches:** It includes transformation-based approach that combines rule-based approach and statistical approach. These approaches helps to achieve a significant improvement of POS performance, since it can combine necessary features from statistical and linguistic based approaches (El Hadj et al., 2009).

Every human language poses its own challenges and requires specific methods. Amharic is also one of the families of morphologically rich languages that has major challenges related to POS tagging task.

All proposed POS taggers were based on ELRC Tagset, developed by different individuals. This paper addresses the various developments in POS-taggers and POS-tagset for the Amharic language, which is very essential computational linguistic tool needed for many NLP applications. We focused on extending existing annotated data (ELRC tag-set), constructing new tag-set and then implementing machine learning methods that have been recently applied to solve POS problems of Amharic Language.

## 2 Previous work

Several attempts have been made in the past to develop POS algorithms for Amharic Language. Some of these works are as follows.

Getachew (2001), attempted to develop a Hidden Markov Model (HMM) based POS tagger using 23 POS tags from 300 words. Since the tag-set and data are very small, the tagger does not have the capability of predicting the POS tag of unknown words.

Adafre (2005), developed a POS tagger using Conditional Random Fields (CRF) and abstract tag-set consisting of 10 tags and obtained overall accuracy of 74% on a manually annotated text corpus of five Amharic news articles (1000 words). The small amount of annotated data leads to drastic impact on tagging accuracy.

Gamback et al. (2009), compared three tagging strategies; HMM, Support Vector Machines (SVM) and Maximum Entropy (ME) using the manually annotated corpus developed by Demeke and Getachew (2006) at the Ethiopian Language Research Center (ELRC) of Addis Ababa University. Since the corpus contained few errors and tagging inconsistencies, they have cleaned the corpus. They obtained the average accuracies (after 10-fold cross validation) of 85.56%, 88.30%, and 87.87% for the HMM, SVM, and ME-based taggers respectively for the ELRC tag-set.

Tachbelie and Menzel (2009), conducted POS tagging experiments for Amharic using uncleaned ELRC corpus in order to use POS information in language modeling. They developed Trigrams'n'Tags (TnT) and SVM-based taggers and compared in terms of performance, tagging speed as well as memory requirements. The results of their experiments show that with respect to accuracy, SVM-based taggers perform better than TnT-based taggers although TnT-based taggers are more efficient with regard to speed and memory requirements. This work lacked a reference allowing for an evaluation of the quality of the annotations that may highly affect the performance of taggers.

Gebre (2010), attempts to improve the performance of Amharic POS tagger based on CRF, SVM, Brill and HMM. Cleaning up ELRC tag-set to minimize the pre-existing tagging errors and inconsistencies can increase the performance of the POS tagger. With 10-fold cross validation they have obtained average accuracy of 90.95%, 90.43%, 87.41%, and 87.09% for CRF, SVM, Brill and TnT taggers respectively. Even though they obtained good accuracy the precision and recall reported in this paper is very far from the average accuracy. For example, by using CRF tagger they have obtained 60% recall and 67% precision and using SVM 64% recall and 68% precision.

## 3 Amharic Language

Amharic is the second most widely spoken Semitic language in the world, after Arabic. It is characterized by complex, productive morphology, with a basic word-formation mechanism, root and pattern (Shashirekha and Gashaw, 2016).
The typical clause order in Amharic is noun + object + verb. Nouns may denote gender, number, definiteness, case, and direct object status by affixes prefixes and suffixes, predominately suffixes. Amharic nouns may have a masculine or feminine gender. Suffixes are added to denote a masculine or feminine noun gender. Some nouns may have both masculine and feminine gender, while other nouns may only have one gender. The feminine gender is used to indicate female as well as the smallness (Degsew, 2014).

## 4 Proposed Approach

Sentence and word tokenization is performed for unannotated Quran and Bible texts before part of tagging process. Since Amharic annotated corpus (ELRC) is limited with only news domain and cleaned version of this dataset is not available, ELRC data is cleaned. The tag-set in ELRC is only 31 tags. It cannot give much information to reliably develop NLP applications. Therefore, ELRC tag-set is extended from 31 to 51 and then new corpus is constructed from Quran and Bible texts including the extended ELRC annotated data. For training and testing algorithms, data should be splited into training and test set, then morphological features are extracted from the training set for CRFSuit tagger only. After-all training and testing are conducted on the three tagging algorithms for each tag-set.

Particularly we applied machine learning approaches for the POS tagging task, and it can be easily interpreted as a classification problem. In this POS task, the limited set of tags are identified from three Corpora and the training examples are the occurrences of the words along with the respective POS category in the context of appearance. A general representation of the POS tagging process is Shown in Figure 1.

We adopt CRF model, which has widely been used in several basic NLP tasks. It is a conditional model that models the conditional probability distribution of tags ($t_1...t_k$) given observation sequences of words ($w_1...w_k$) in the sentences i.e. $P(t...t_k|w_1...w_k)$. The probability of transition between tags is depends on the previous and next observations. This enables reasoning based on wide contexts, which seems especially important in the case of POS tagging tasks. For large and structured tag-sets, CRF work well with many features that may be mutually dependent (Lafferty et al., 2001).

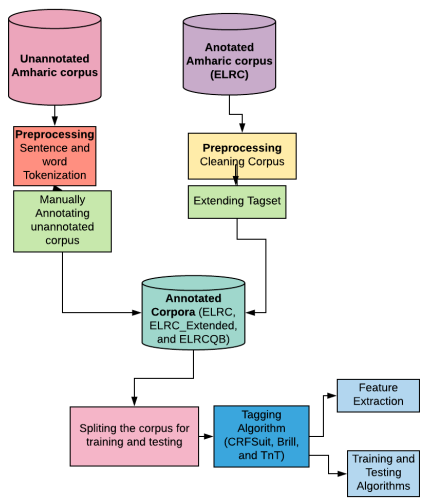

Figure 1: General framework of the proposed approach

Linear−chain CRF is the most popular class of CRFsuit suitable for tagging. CRFSuit training consists of estimation of weight values. A high weight value indicates that strong evidence has been found to support the relation between observations and tags as expressed by the feature (Radziszewski, 2013). Tagging with a trained CRFsuit consists in finding a tag sequence that maximizes the conditional probability. The optimization algorithms used in these work is the Limited-memory Broyden-Fletcher-Goldfarb-Shanno (L-BFGS) algorithm (Saputro and Widyaningsih, 2017) which is employed for solving high-dimensional minimization problems in scenarios where both the objective function and its gradient can be computed analytically. L-BFGS algorithm stores information about the spatial displacement and the change in gradient and uses them to estimate a search direction without storing or computing the Hessian explicitly (Coppola and Stewart, 2014).

### 4.1 Feature extraction

Since Amharic is morphologically rich language, there are a lot of morphological features which enables the POS tagger to predict correctly. In a CRFSuit, each feature function is a function that takes in as input: a sentence s, the position i of a word in the sentence, either the word comes first/last, has hyphen, is current word/previous word/next word digit, alphanumeric, prefix-1, prefix-2, prefix-3, suffix-1, suffix-2, suffix-3, Previous-1 word tag, and Previous-2 word tag. For each such value configuration, a separate function must be provided in advance. In order to train and test the POS tagger, we define a function that can extract all the above features and then used to input in CRFSuit feature function.

### 4.2 Dataset description

The dataset used in this study are categorized in to three, ELRC annotated corpus that contains 210,000 words (Demeke and Getachew, 2006), extended re-tagged corpus of ELRC, and the newly annotated corpus of the Amharic translation of Quran, and Bible.
In the first domain, the tag-set is based on 11 basic tags, most of which have further been refined to provide more linguistic information, thus increasing the tag-set to 31.
Even though, (Gebre, 2010) cleaned ELRC tagged corpus, we couldn't get the cleaned one. Therefore, we have enforced to clean again by following the strategies used to clean in this work.
Since Amharic is morphologically complex language, 31 tags of ELRC tag-set cannot give much information to reliably develop NLP applications. Some tags that may be critical depending on the target application are missing. In detail, the limitation of ELRC tag-set is reported by (Gebre, 2010). Furthermore, we extended ELRC tag-set from 31 tags to 51 tags by adding S for those tags with plural numbers such as Noun with plural numbers (NS) and the addition of preposition and conjunction with adverbs then we called this dataset ELRC-Extended. The third category is the extended ELRC tagset plus manually tagged Quran and Bible Documents by taking ELRC as a base, we call this new dataset ELRCQB which contains 62 tags. ELRCQB dataset size is 33,940 sentences (440,941 words). For Example the distribution of ELRC-Extended tag-set is shown in figure 2.

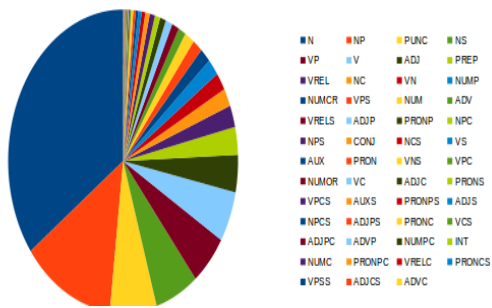

Figure 2: ELRCExtended tag-set distribution

## 5 Experiments and Results

This section presents the experiments validating the three machine learning algorithms (Brill, TnT, and CRFSuit Taggers) implemented for Amharic POS Tagging task. We use sklearn-crfsuite which is a CRFsuite (python-crfsuite) wrapper that provides scikit-learn compatible sklearn-crfsuite.CRF estimator to train and test our POS tagger. 10-Fold cross-validation is applied for training and then evaluating all tagging techniques for all three corpora. 10-fold cross-validation data for all tagsets is shown on table 2 with the information of known and unknown words of the testing data for each fold.

The results obtained by applying the three different tagging strategies are shown in Table 2. TnT tagger and Brill tagger performs almost the same. CRFSuit Tagger achieves the best scores of all three taggers. Because the features extracted from the tag-sets enables the system to predict the words tag even if it is not in training data. To handle unknown words for Brill and TnT tagger, we used n-gram tagger as back-of tagging strategies that assign a maximum tag appeared in the test set.

All approaches are evaluated using confusion matrix. In this POS tagging problem, the confusion matrix contains 62 rows and 62 columns, 52 rows and 52 columns, 31 rows and 31 columns for ELRCQB, ELRC-Extended and ELRC tag-sets of all tagger. But due to a large number of tags that may not have good visibility, we showed only the first top 20 tags of confusion matrix for the best score. Confusion matrix helps to indicate correctly classified and wrongly classified elements of each class. For example, in figure 5 the proposed approach (CRFSuit Tagger), the tag N, 9388 of 9930 are classified correctly but the remaining 542 are misclassified as different tags. The vertical lines associated with each confusion matrix indicates the elements in the class with a maximum number of elements predicted.

The base for our work is (Gebre, 2010), that yields good performance for different machine learning approaches using 10 fold cross-validation technique, which is the overall accuracy of 90.95, 90.43, 87.41, 87.09 for CRF, SVM, Brill, and TnT taggers respectively on cleaned ELRC tagset only. Even though overall accuracy is reported as above the precision and recall result reported in his work is less compared to our precision and recall results showed in table 3.

The main contribution of this work is extending ELRC tagset, constructing new tagset from Quran and Bible document and parameter tuning by selecting the best parameter of Sklearn-CRFSuit through grid searching which is C1:0.064 and C2:0.002. We achieved the overall average accuracy of 86.44, 95.87, and 92.27 for ELRC, ELEC-Extended and ELRCQB tagsets respectively. As the result indicated extending the tag-set increased the performance by 9.43 which is very significant. While the domain of the tag-set increasing the number of unknown words also increased and the style of writing in Quran and Bible is different, then it creates misclassification. Even though increasing the tag-set size has its own advantage in machine learning approaches in general, in this work the noise from domain difference creates 3.6% performance difference between ELRC-Extended and ELRQB tag-sets.

## 6 Conclusion

In this paper, we have described three machine learning approaches for automatic tagging of Amharic text processing. The models described here are very simple and efficient for automatic tagging. The extended and newly constructed tag-sets have contributed to the high performance of our proposed approach and it will contribute to the reliable development of applications of machine translation, information retrieval, information extraction and speech synthesis/recognition. The best performances are achieved for the CRFSuit learning model along with the important morphological features extracted from the training set.

Table 1: 10-Fold Cross Validation Data

| Fold | # ELRC Words | | | | # ELRC_Extended Words | | | | # ELRCQB Words | | | |
|---|---|---|---|---|---|---|---|---|---|---|---|---|
| | Training | Testing | | | Training | Testing | | | Training | Testing | | |
| | | Known | Unknown | Total | | Known | Unknown | Total | | Known | Unknown | Total |
| 1 | 180788 | 18102 | 2744 | 20846 | 185091 | 18433 | 2646 | 21079 | 411643 | 27147 | 4245 | 31392 |
| 2 | 181435 | 17637 | 2562 | 20199 | 185447 | 18233 | 2490 | 20723 | 413002 | 26065 | 3968 | 30033 |
| 3 | 182222 | 17155 | 2257 | 19412 | 186381 | 17634 | 2155 | 19789 | 414879 | 24716 | 3440 | 28156 |
| 4 | 181722 | 17532 | 2380 | 19912 | 185532 | 18344 | 2294 | 20638 | 416764 | 22819 | 3452 | 26271 |
| 5 | 180956 | 18188 | 2490 | 20678 | 185089 | 18659 | 2422 | 21081 | 376787 | 57808 | 8440 | 66248 |
| 6 | 182679 | 16867 | 2088 | 18955 | 186892 | 17233 | 2045 | 19278 | 370077 | 64255 | 8703 | 72958 |
| 7 | 181368 | 17841 | 2425 | 20266 | 185415 | 18409 | 2346 | 20755 | 370601 | 63842 | 8592 | 72434 |
| 8 | 181189 | 17965 | 2480 | 20445 | 185100 | 18692 | 2378 | 21070 | 407013 | 30114 | 5908 | 36022 |
| 9 | 181735 | 17588 | 2311 | 19899 | 185747 | 18208 | 2215 | 20423 | 405709 | 31495 | 5831 | 37326 |
| 10 | 180612 | 18639 | 2383 | 21022 | 184836 | 19000 | 2334 | 21334 | 400840 | 33654 | 8541 | 42195 |
| Average | 181470.6 | 17751.4 | 2412 | 20163.4 | 185553 | 18284.5 | 2332.5 | 20617 | 398731.5 | 38191.5 | 6112 | 44303.5 |

Table 2: Average 10-Fold Accuracy of Brill, TnT and CRFSuit Taggers

| Tagger | ELRC | | | ELRC-Extended | | | ELRCQB | | |
|---|---|---|---|---|---|---|---|---|---|
| | Known words | Unknown words | Overall | Known words | Unknown words | Overall | Known words | Unknown words | Overall |
| Brill | 89.185 | 25.97 | 81.627 | 99.997 | 46.206 | 93.876 | 97.359 | 33.409 | 88.539 |
| TnT | 89.889 | 25.969 | 82.25 | 99.997 | 46.214 | 93.877 | 97.491 | 33.409 | 88.661 |
| CRFSuit | 87.883 | 75.79 | 86.442 | 99.069 | 70.872 | 95.868 | 96.408 | 65.634 | 92.242 |

Table 3: CRFSuit Tagger Best score of Average precision, recall and f-1score on ELRC, ELRC-Extended, and ELRCQB tag-set

| Tag-set | precision | recall | f1-score | support |
|---|---|---|---|---|
| ELRC | 0.902 | 0.898 | 0.899 | 20445 |
| ELRC-Extended | 0.97 | 0.97 | 0.97 | 21070 |
| ELRCQB | 0.951 | 0.951 | 0.951 | 36022 |

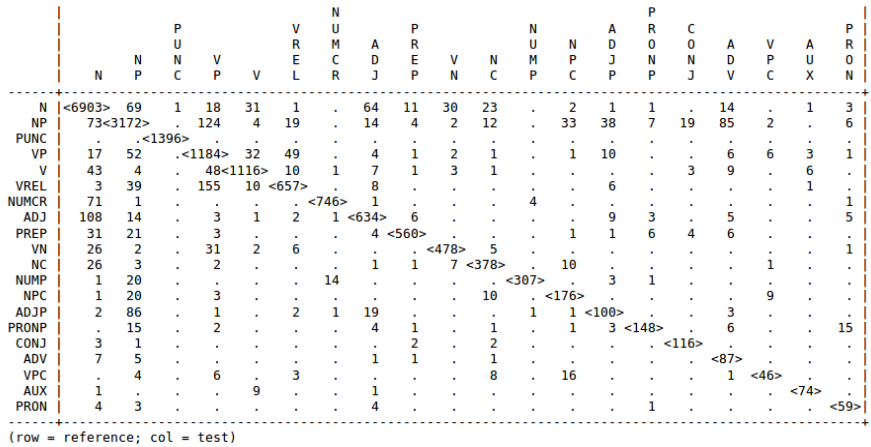

Figure 3: Confusion matrix of Top 20 Tags best score for ELRC tagset

All the tag-sets we have used in this work lacks expert knowledge. Therefore, it should be standardized to obtain enhanced performance. It is very limited to identify names of people and places, which is critical for information extraction. The presence of a proper noun tag is even more important in the context of Amharic, but the idea of letter case distinction does not exist and most Ethiopian names are just normal words in the language. Thus, most proper nouns that are easily recognized in English by the case of the initial letter cannot be recognized in Amharic.